\documentclass[letterpaper]{article} 
\usepackage{aaai2026}  
\usepackage{times}  
\usepackage{helvet}  
\usepackage{courier}  
\usepackage[hyphens]{url}  
\usepackage{graphicx} 
\usepackage{natbib}  
\usepackage{caption} 
\usepackage{algorithm}
\usepackage{algorithmic}
\usepackage{booktabs} 

\usepackage{newfloat}
\usepackage{listings}
\DeclareCaptionStyle{ruled}{labelfont=normalfont,labelsep=colon,strut=off} 
\floatstyle{ruled}
\newfloat{listing}{tb}{lst}{}
\floatname{listing}{Listing}
\title{Quantifying Conversational Reliability of Large Language Models under Multi-Turn Interaction}
\author{
    Jiyoon Myung
}
\affiliations{
    Samsung SDS\\
    jiyoon0424@gmail.com
}

\usepackage{bibentry}

\begin{document}

\maketitle

\begin{abstract}
Large Language Models (LLMs) are increasingly deployed in real-world applications where users engage in extended, mixed-topic conversations that depend on prior context. Yet, their reliability under realistic multi-turn interactions remains poorly understood. We conduct a systematic evaluation of conversational reliability through three representative tasks that reflect practical interaction challenges: (1) maintaining global constraints across topic shifts, (2) selecting the correct tool or agent amid interleaved intents, and (3) tracking structured entities under revisions and distractions. Each task pairs single-turn and multi-turn settings, allowing us to quantify reliability degradation under extended dialogue. Across both commercial and open-source models, we observe substantial declines in reliability, particularly for smaller models. Error analyses reveal recurring failure modes such as instruction drift, intent confusion, and contextual overwriting, which compromise dependable behavior in operational systems. Our findings highlight the need for stress-testing LLMs for conversational reliability and developing more robust evaluation methods for trustworthy deployment.
\end{abstract}

\section{Introduction}

Deployed conversational systems must operate reliably in messy, multi-turn environments: users shift topics, interleave irrelevant content, and revise their goals mid-dialogue. Models are therefore expected to remain consistent and robust as conversations grow longer and more context-dependent. Failures in these basic interactive abilities can seriously undermine the usability of a system.  

Empirical studies show that large language models (LLMs) often struggle under such conditions. Multi-turn analyses reveal substantial degradation in reliability compared to single-turn prompts~\cite{llmslost}, while long-context evaluations expose weaknesses such as the ``lost in the middle'' effect~\cite{lostmiddle}. Several benchmarks---such as Multi-IF~\cite{multiif}, StructFlowBench~\cite{structflowbench}, MMMT-IF~\cite{mmmtif}, and MINT~\cite{mint}---probe aspects of conversational robustness, including instruction consistency and tool use, but they often focus on abstract challenges or rely on subjective judgments. This leaves open the question of how to objectively evaluate concrete behaviors required in practice.  

We address this gap by introducing three compact, pass/fail evaluable tasks that reflect core requirements for real-world assistants:
\begin{itemize}
    \item \textbf{Instruction Following under extended conversation}: enforcing a global style constraint despite distracting turns.  
    \item \textbf{Tool Selection in mixed-topic dialogues}: routing each request to the correct tool when multiple intents are interleaved.  
    \item \textbf{Entity Extraction under revisions and distractions}: tracking the user’s final structured intent despite changes of mind, chit-chat, or irrelevant mentions.  
\end{itemize}

Each task is paired with a single-turn counterpart, allowing us to isolate and quantify the degree of \emph{reliability degradation} that occurs under extended, mixed-topic interactions.  
This paired design enables reproducible and objective assessment of how conversational reliability deteriorates across model families and dialogue lengths.  
By comparing large commercial LLMs with smaller open-source counterparts, we further reveal capacity-dependent vulnerabilities that have direct implications for real-world deployment and safety.  

In summary, our study bridges the gap between research benchmarks and practical evaluation by providing deterministic, reproducible tests of conversational robustness—highlighting where current models fail to sustain reliable behavior over time.

\begin{figure*}[t]
    \centering
    \includegraphics[width=\textwidth]{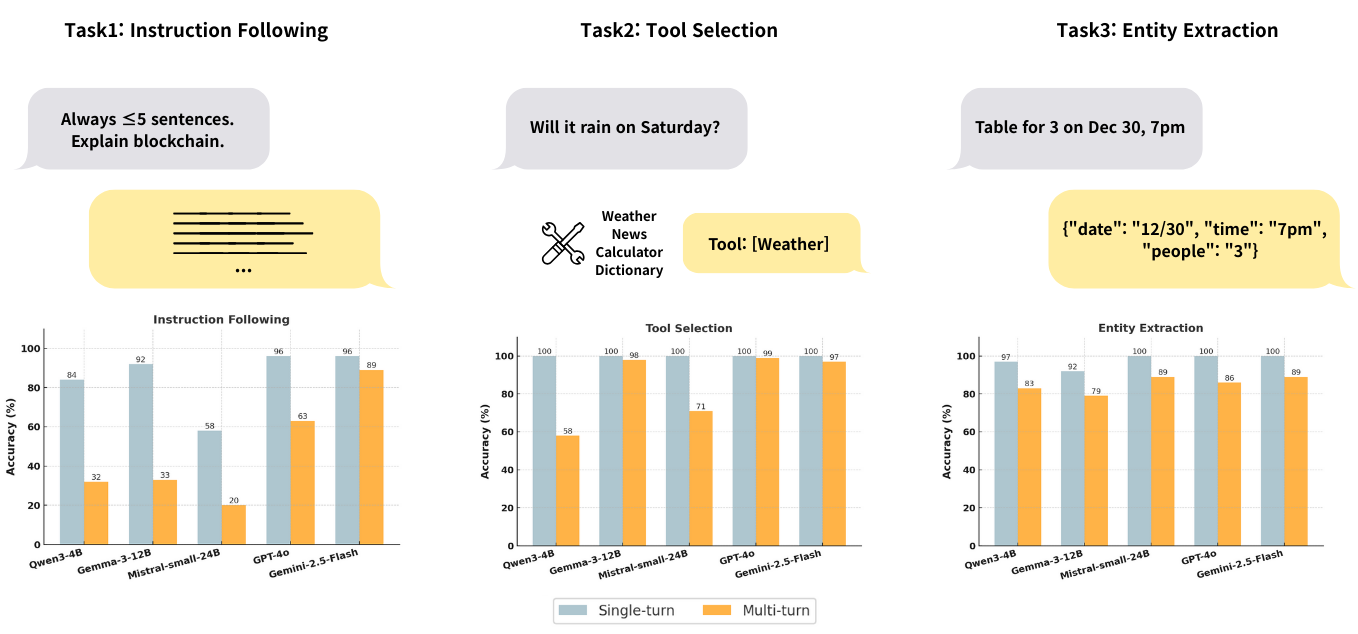}
    \caption{
    Single-turn vs Multi-turn accuracy across three evaluation tasks. 
    Each panel shows a task example on the left and model accuracy on the right. 
    Performance drops most severely in \textbf{Instruction Following}, 
    while \textbf{Entity Extraction} remains relatively robust, and 
    \textbf{Tool Selection} shows mixed degradation depending on model size.
    }
    \label{fig:multiturn_tasks}
\end{figure*}

\section{Experiments}

Our goal is to quantify how performance degrades when tasks are embedded in extended, mixed-topic conversations, compared to their single-turn counterparts. 
We therefore define three representative tasks motivated by real-world service scenarios, construct synthetic single-turn and multi-turn dialogues, and evaluate a range of models from commercial LLMs to smaller open-source SLMs. 
We report task accuracy and analyze common error patterns, highlighting the operational risks that arise in multi-turn interactions.

\subsection{Task Scenarios}
We focus on three multi-turn challenges that frequently arise in real-world conversational systems:

\textbf{Instruction Following.}  
A length constraint is specified at the beginning of the dialogue (e.g., ``always answer in at most 5 sentences''). 
The conversation then continues with unrelated topics over multiple turns, and the final user request is deliberately phrased to elicit a long and detailed answer.  
The evaluation measures whether the model consistently respects the global style constraint 
throughout the dialogue, even when pressured to violate it.  
This represents scenarios where chatbots must reliably enforce prescribed formatting rules, 
such as staying concise or avoiding certain tokens.

\textbf{Tool Selection.}  
At specific turns, the model must select the correct tool from a fixed set: 
\texttt{[Weather, News, Calculator, Stock, Recipe, Dictionary]}.  
In the single-turn setting, the user query directly corresponds to a single tool.  
In the multi-turn setting, conversations are explicitly constructed to \emph{mix multiple topics} from the tool list, 
so that the model must correctly route each request in a more challenging environment.  
This reflects real-world situations such as intent classification in digital assistants 
or routing queries to the right component in multi-agent systems.

\textbf{Entity Extraction.}  
The task is to extract the final structured slots in a restaurant reservation scenario: \texttt{(date, time, number of people)}.  
In single-turn cases, the reservation request is stated directly.  
In multi-turn cases, we introduce realistic complications: users may change their mind multiple times (\emph{change in mind}), engage in unrelated small talk (\emph{intermediate chit chat}), or mention other people's reservations (\emph{multiple mention}).  
The model must correctly track the user’s final intent and output the canonical reservation values.  
This setting mirrors practical needs such as extracting parameters for executing tools (e.g., calendar or booking APIs) 
or supporting real agents that must handle evolving goals in natural conversations.

\subsection{Data Generation}
For each task, we generated paired single-turn and multi-turn dialogue datasets.

\begin{itemize}
    \item \textbf{Dialogue generation}: We used GPT-5 to synthesize dialogues, controlling for dialogue length, number of topic shifts, and frequency of modifications. Synthetic generation allowed us to systematically vary factors such as distraction density or revision frequency while keeping overall style consistent. 
    \item \textbf{Single-turn vs. multi-turn}: Each instance was created in two variants—one where the relevant request was posed directly (single-turn), and one where it was embedded within a longer conversation with distractions, topic changes, or corrections (multi-turn).
    \item \textbf{Task-specific constraints}:  
    For \emph{Instruction Following}, dialogues contained between 5 and 15 turns. This range was chosen to roughly approximate short-to-moderate real customer service sessions, where most exchanges end within a dozen turns.  
    For \emph{Tool Selection}, dialogues ranged from 6 to 16 turns, and the number of distinct relevant tools was randomized between 2 and 6 to mimic real assistants that must juggle several intents in a single session.  
    For \emph{Entity Extraction}, each dialogue randomly combined conditions such as \emph{change in mind}, \emph{intermediate chit chat}, and \emph{multiple mention}, reflecting diverse patterns observed in reservation-style conversations. 
    \item \textbf{Size}: For each task, we generated approximately 100 dialogues per condition, yielding about 600 evaluation cases in total across all tasks.
    \item \textbf{Annotation}: Ground-truth labels were automatically derived during generation (e.g., correct tool, final reservation details) and verified by human inspection on a sample basis. 
    \item \textbf{Availability}: The generated datasets will be released on HuggingFace after publication to support reproducibility and further research.
\end{itemize}

\subsection{Models Evaluated}
We evaluated a diverse set of language models covering both commercial and open-source deployments. All models were accessed via their respective official APIs, and we fixed the decoding temperature to $0$ to ensure deterministic outputs.
\begin{itemize}
    \item \textbf{Commercial LLMs}: GPT-4o, GPT-4o-mini, Gemini-2.5-Flash.
    \item \textbf{Open-source SLMs}: Qwen-8B, Qwen-32B, Ministral-8B, Mistral-small-24B, Gemma-3-12B.
\end{itemize}

\subsection{Evaluation Metrics}
As all tasks were designed with clear pass/fail criteria, we adopt \emph{accuracy} as the primary metric to ensure clarity, replicability, and ease of interpretation across diverse models. Unlike open-ended generation or preference-based evaluation, our tasks do not benefit from graded or subjective metrics.

\begin{itemize}
    \item \textbf{Instruction Following}: An output is correct if it satisfies the constraint of containing at most five sentences. Any response exceeding this length limit is considered incorrect.
    \item \textbf{Tool Selection}: An output is correct if the tool selected by the model matches the ground-truth tool for that turn.
    \item \textbf{Entity Extraction}: An output is correct if the extracted \texttt{(date, time, number of people)} exactly matches the ground-truth values provided with the dialogue.
\end{itemize}

\section{Results}

\begin{table*}[t]
\centering
\small
\begin{tabular}{lcccc}
\toprule
& \textbf{GPT-4o} & \textbf{GPT-4o-mini} & \textbf{Gemini-2.5-Flash} & \textbf{Gemma-3-12B} \\
\midrule
\multicolumn{5}{l}{\textit{Single-turn}} \\
Instruction Following & 96 & 93 & 96 & 92 \\
Tool Selection        & 100 & 100 & 100 & 100 \\
Entity Extraction     & 100 & 96 & 100 & 92 \\
\midrule
\multicolumn{5}{l}{\textit{Multi-turn}} \\
Instruction Following & 63 & 24 & 89 & 33 \\
Tool Selection        & 99 & 93 & 97 & 98 \\
Entity Extraction     & 86 & 84 & 89 & 79 \\
\bottomrule
\end{tabular}

\vspace{0.3cm} 

\begin{tabular}{lccccc}
\toprule
& \textbf{Qwen3-4B} & \textbf{Qwen3-8B} & \textbf{Qwen3-32B} & \textbf{Ministral-8B} &  \textbf{Mistral-small-24B} \\
\midrule
\multicolumn{5}{l}{\textit{Single-turn}} \\
Instruction Following & 84 & 83 & 92 & 27 & 58 \\
Tool Selection        & 100 & 100 & 100 & 99 & 100 \\
Entity Extraction     & 97 & 98 & 100 & 100 & 100 \\
\midrule
\multicolumn{5}{l}{\textit{Multi-turn}} \\
Instruction Following & 32 & 27 & 54 & 11 & 20 \\
Tool Selection        & 58 & 89 & 47 & 37 & 71 \\
Entity Extraction     & 83 & 88 & 89 & 88 & 89 \\
\bottomrule
\end{tabular}
\caption{Accuracy (\%) of all models across tasks in single-turn and multi-turn settings.}
\label{tab:main_results}
\end{table*}

\subsection{Overall Results}

Table~\ref{tab:main_results} summarizes the main experimental results.  
Across all tasks, we observe a consistent degradation when moving from single-turn to multi-turn settings.  
McNemar tests confirm that these performance gaps are statistically significant across all three tasks (see Appendix for full results).  
This confirms that multi-turn dialogue introduces substantial additional difficulty across model scales.

\textbf{Instruction following} exhibited the largest degradation overall. 
While most models performed strongly in the single-turn setting, accuracy dropped sharply once the task was embedded in multi-turn dialogues. 
This decline was evident even for commercial LLMs (e.g., GPT-4o falling from 96\% to 63\%, Gemini-2.5-Flash from 96\% to 89\%), and was even more severe for smaller models (e.g., GPT-4o-mini at 24\%, Qwen3-8B at 27\%). 
These results suggest that maintaining global constraints over extended conversations remains a fundamental challenge, regardless of model capacity.

\textbf{Tool selection}, by contrast, showed strong robustness among commercial models. 
Systems such as GPT-4o and Gemini-2.5-Flash  maintained very high accuracy ($\geq$97\%) even in multi-turn settings. 
However, smaller open-source models displayed substantial degradation, particularly when multiple tools were relevant in the same dialogue. 
For example, Qwen3-32B dropped to 47\% and Ministral-8B to 37\%. 
This indicates that explicit grounding to a fixed tool set is relatively easy for larger models, but smaller ones struggle to maintain consistency under higher conversational complexity.

\textbf{Entity extraction} emerged as the most robust task across models. 
In single-turn reservations, nearly all models achieved near-perfect accuracy (96–100\%). 
Even in multi-turn dialogues, where user changes of mind and distractions were introduced, 
performance remained relatively high (typically 84–89\%, with the lowest at 79\%). 
This resilience likely stems from the structured nature of the target fields---\texttt{(date, time, number of people)}---which are expressed as explicit numbers or short phrases. 
As a result, models could reliably capture the final slot values with limited ambiguity. 
We hypothesize that if the extraction targets had required richer contextual understanding, 
for example, mapping free-form user mentions like “the pizza with pineapple” to a canonical menu item 
(\textit{Hawaiian pizza}), the degradation would have been much more pronounced.

We also note model-specific patterns. \textbf{Mistral} was particularly weak at the instruction-following task, even in single-turn scenarios (27-58\%), suggesting difficulty in adhering to length constraints. On the other hand, \textbf{Gemma} models demonstrated remarkable robustness in tool selection, with multi-turn performance staying almost at the same level as single-turn. Interestingly, larger models (e.g., GPT-4o, Gemma-3-27B, Qwen-32B) showed smaller performance gaps between single-turn and multi-turn than smaller ones (e.g., GPT-4o-mini, Mistral-8B), underscoring the role of model capacity in sustaining performance under long-horizon interactions.  

Overall, these findings reinforce that while multi-turn dialogue universally degrades accuracy, the degree of impact depends strongly on the task type and model family, with global instruction maintenance emerging as the most challenging dimension.

\subsection{Detailed Error Analysis}
\label{sec:error_analysis}

To understand the causes of reliability degradation, we analyze results by conversation length, task complexity, and entity extraction scenario type (see Appendix for full tables).

Conversation length alone is not a dominant factor for instruction-following degradation. Accuracy fluctuates but does not consistently worsen in longer dialogues; at 10 turns it even peaks at 96\%. This suggests that failures arise not merely from context length but from specific distractors or competing user demands.

Task complexity shows clearer effects: accuracy in tool selection declines sharply as the number of candidate tools increases—from 98\% with two tools to 64\% with five. Models struggle to identify relevant context amid multiple distractors, implying tangible risks for real multi-agent or tool-use systems.

Entity extraction errors vary by scenario. The \emph{date} slot is consistently weakest, reflecting difficulty in temporal tracking. \emph{Change in mind} conversations are most error-prone (85\%), while \emph{multiple mention} cases are relatively robust (91\%). Conversational distractions such as temporal shifts or irrelevant chatter differentially impact reliability in realistic reservation tasks.

\subsection{Qualitative Error Analysis}
\label{sec:qualitative}

Representative qualitative examples (see Appendix for full results) illustrate the characteristic failure modes in multi-turn settings. These cases reveal how long, information-heavy prompts, topic shifts, and misleading mentions break conversational consistency and gradually erode task reliability.

In the \emph{Instruction Following} case, the model ignores the global constraint after several irrelevant turns—producing a thirteen-sentence historical summary despite being instructed to stay within five. This shows a gradual loss of constraint adherence as dialogues extend, indicating that even stylistic or formatting constraints are fragile under sustained interaction.  
In the \emph{Tool Selection} task, topic mixing causes the model to reuse the previous tool (“Stock”) even when the next user query clearly requires a different one (“Weather”), reflecting overcommitment to recent context and insufficient intent re-evaluation. Such behavior can compound in realistic assistants that must switch between domains dynamically.  
For \emph{Entity Extraction}, models correctly update some slots but are distracted by nearby mentions, overwriting the final reservation time. This illustrates how transient context interference disrupts working memory and undermines the reliability of structured information tracking over dialogue turns.

Together, these qualitative results confirm that degradation arises not from length alone but from specific context conflicts and memory overwriting across turns.

\section{Conclusion}

We investigated how conversational reliability degrades when tasks are embedded in extended, mixed-topic dialogues.  
To capture this effect in a controlled and reproducible way, we introduced three compact multi-turn evaluation tasks—global instruction following, tool selection, and reservation entity extraction—each paired with a single-turn counterpart.  
This design allowed us to isolate the degree of reliability degradation and quantify how conversational context affects model stability across both large commercial LLMs and smaller open-source SLMs.  
We found that models which behave reliably in simple settings often show sharp declines in consistency when conversations become longer and more dynamic, with smaller models especially vulnerable.

Our contribution is to ground multi-turn evaluation in scenarios practitioners actually face when deploying conversational systems.  
Unlike abstract benchmarks, our tasks emphasize practical reliability factors—sustaining consistent behavior, routing diverse requests, and maintaining structured state over time.  
The evaluation framework yields reproducible pass/fail metrics and task-specific error analyses that provide actionable insights into where conversational fragility emerges.  
These findings highlight the need to treat multi-turn reliability as a core dimension of model evaluation and to develop methods that better preserve robustness under realistic dialogue conditions.


\bibliography{main}

\newpage
\appendix

\section{Detailed Quantitative Results}
\label{appendix:analysis_tables}

This appendix provides detailed quantitative tables referenced in Detailed Error Analysis Section.
The results include breakdowns by conversation length, number of tools, and entity extraction scenario type.

\begin{table}[h]
\centering
\caption{Instruction Following accuracy by number of dialogue turns.}
\label{tab:instr_length}
\begin{tabular}{lcc}
\toprule
\textbf{Turns} & \textbf{Accuracy} \\
\midrule
5 & 0.40 \\
6 & 0.28 \\
7 & 0.38 \\
8 & 0.15 \\
9 & 0.29 \\
11 & 0.25 \\
\bottomrule
\end{tabular}
\end{table}

\begin{table}[h]
\centering
\caption{Tool Selection accuracy by dialogue length.}
\label{tab:tool_length}
\begin{tabular}{lcc}
\toprule
\textbf{Turns} & \textbf{Accuracy} \\
\midrule
6 & 0.76 \\
7 & 0.74 \\
8 & 0.82 \\
9 & 0.74 \\
10 & 0.96 \\
11 & 0.78 \\
12 & 0.80 \\
13 & 0.75 \\
14 & 0.81 \\
15 & 0.75 \\
\bottomrule
\end{tabular}
\end{table}

\begin{table}[h]
\centering
\caption{Tool Selection accuracy by number of available tools.}
\label{tab:tools}
\begin{tabular}{lcc}
\toprule
\textbf{\# Tools} & \textbf{Accuracy} \\
\midrule
2 & 0.98 \\
3 & 0.82 \\
4 & 0.74 \\
5 & 0.64 \\
6 & 0.71 \\
\bottomrule
\end{tabular}
\end{table}

\onecolumn

\begin{table}[h]
\centering
\caption{Entity Extraction accuracy by scenario type.}
\label{tab:entity_type}
\begin{tabular}{lcccc}
\toprule
\textbf{Scenario Type} & \textbf{Date} & \textbf{Time} & \textbf{People} & \textbf{All Match} \\
\midrule
Change in Mind & 0.91 & 0.96 & 0.99 & 0.85 \\
Intermediate Chit Chat & 0.91 & 0.97 & 0.97 & 0.86 \\
Multiple Mention & 0.94 & 0.98 & 0.99 & 0.91 \\
\bottomrule
\end{tabular}
\end{table}

\section{Representative Qualitative Examples}
\label{appendix:qualitative}

Table~\ref{tab:qual_errors} summarizes qualitative error examples referenced in Qualitative Error Analysis Section.
These cases illustrate the major failure modes—such as loss of instruction adherence, intent confusion, and distraction by irrelevant mentions—that contribute to the observed reliability degradation.

\begin{table}[h]
\centering
\caption{Representative qualitative error cases.}
\label{tab:qual_errors}
\begin{tabular}{p{6cm} p{5cm} p{5cm}}
\toprule
\textbf{Context (excerpt)} & \textbf{Model Output} & \textbf{Ground Truth}  \\
\midrule
\textbf{Instruction Following} \newline
User: From now on, always answer in at most 5 sentences. \newline
(after 10 turns of unrelated chit-chat) \newline
User: Could you walk me through the entire rise and fall of the Byzantine Empire?
&
Provides a long, detailed explanation in \textcolor{red}{13} sentences. 
&
Concise 4-5 sentence summary.  \\
\midrule
\textbf{Tool Selection} \newline
User: What is the stock price of AAPL today? \newline
User: And what is the weather in Paris right now?
&
\textcolor{red}{\texttt{[Stock]}}
&
\texttt{[Weather]}  \\
\midrule
\textbf{Entity Extraction} \newline
User: I'd like a table for 3 on April 18 at 7 pm. \newline
User: By the way, my friend Sarah booked for 6pm yesterday. \newline
User: Oh actually, make it 4 people. \newline
&
\{"date": "04/18", "time": "\textcolor{red}{7pm}", "people": "4"\}
&
\{"date": "04/18", "time": "6pm", "people": "4"\} \\
\bottomrule
\end{tabular}
\caption{Representative qualitative error cases. Models frequently (i) lose track of global instructions, (ii) select the wrong tool when topics shift, or (iii) are distracted by irrelevant mentions.}
\label{tab:qual_errors}
\end{table}

\section{Statistical Significance Tests}
\label{appendix:mcnemar}

To verify that the observed performance gaps between single-turn and multi-turn settings are not due to random variation, 
we conducted McNemar tests for each task. 
Both the exact binomial version and the continuity-corrected $\chi^2$ approximation were applied 
to paired single-turn vs.\ multi-turn samples ($n=900$ per task).  

\begin{table*}[h]
\centering
\small
\begin{tabular}{lccccc}
\toprule
\textbf{Task} & \textbf{Single-turn Acc.} & \textbf{Multi-turn Acc.} & \textbf{Exact $p$} & \textbf{$\chi^2$ (cc)} & \textbf{$p$} \\
\midrule
Instruction Following & 0.764 & 0.396 & $1.7\times 10^{-61}$ & 244.56 & $3.99\times 10^{-55}$ \\
Tool Selection        & 0.999 & 0.763 & $3.0\times 10^{-64}$ & 210.00 & $1.37\times 10^{-47}$ \\
Entity Extraction     & 0.981 & 0.924 & $5.2\times 10^{-9}$  & 31.65  & $1.85\times 10^{-8}$  \\
\bottomrule
\end{tabular}
\caption{McNemar test results on 900 paired samples per task. 
All differences between single-turn and multi-turn settings are statistically significant.}
\label{tab:mcnemar}
\end{table*}

\section{Prompt Templates for Data Generation}
\label{appendix:prompts}

This appendix lists the full prompts used to generate synthetic dialogues 
for the three evaluation tasks. 
We provide separate templates for single-turn and multi-turn settings. 
For reproducibility, all experiments were generated using GPT-5 with controlled random seeds. 
In addition, dialogue length, topic composition, and distraction density were parameterized 
through format variables, allowing us to systematically control the experimental conditions 
while maintaining consistent prompt structure.

\subsection{Instruction Following}
\textbf{Single-turn prompt.}
\begin{lstlisting}
You are a data generator for a single-turn instruction-following benchmark.

Goal:
Return ONLY a JSON object that contains exactly ONE user message (no assistant message).

Rules:
- Language: English.
- The single user message MUST begin with this instruction:
  "Always answer in up to 5 sentences."
- In the SAME user message, after the instruction, ask a natural question about the topic: "{topic}".
- The question should naturally tempts the assistant to give a long, detailed answer (e.g., asking for a full story summary, explanation of a complex topic, or step-by-step process).
- Do NOT include any assistant message.
- Output ONLY a valid JSON object with this exact structure:

{{
  "messages": [
    {{"role": "user", "content": "<single user message that includes the instruction and the question>"}}
  ]
}}
\end{lstlisting}
\vspace{1em}
\textbf{Multi-turn prompt.}
\begin{lstlisting}
You are a data generator for a multi-turn conversation benchmark.

Rules:
- The conversation must be in English.
- The first user message must be: "From now on, always answer in at most 5 sentences"
- The conversation should have {turn} turns total, alternating between user and assistant.
- Make the natural conversation about {topic} after the first instruction.
- The assistant must follow the instruction strictly (at most 5 sentences per reply).
- The last turn must be from the user.
- The last user message should naturally tempt the assistant to give a long, detailed answer (e.g., "Can you tell me the whole story of Cinderella in detail?" or "Explain the entire process of how airplanes are built").
- Return ONLY a valid JSON object in this format:
{{
    "messages": [
        {{"role": "user", "content": "From now on, always answer in at most 5 sentences."}},
        {{"role": "assistant", "content": "<assistant message 1, at most 5 sentences>"}},
        {{"role": "user", "content": "<user message 2>"}},
        {{"role": "assistant", "content": "<assistant message 2, at most 5 sentences>"}},
        ...
        {{"role": "user", "content": "<final user message tempting a long answer>"}}
    ]
}}
\end{lstlisting}

\subsection{Tool Selection}
\textbf{Single-turn prompt.}
\begin{lstlisting}
You are a data generator for a multi-turn tool selection benchmark.

Available tools:
{tools}

Your task:
- Generate exactly ONE user message in JSON format (no assistant replies).
- The single user message should be a natural request that clearly requires exactly ONE of the tools from the list.
- The message should be natural and potentially ambiguous in style, but must still be clear enough for a human to choose the correct tool without additional context.
- The "answer" field should contain ONLY the correct tool name for that user message.
- The output must strictly follow this JSON format:

{{
    "messages": [
        {{"role": "user", "content": "<single user request>"}}
    ],
    "answer": "<one of the tools>"
}}

Do NOT include any explanations, only return the JSON object.
\end{lstlisting}
\vspace{1em}
\textbf{Multi-turn prompt.}
\begin{lstlisting}
You are a data generator for a multi-turn tool selection benchmark.

Available tools:
{tools}

Your task:
- Generate exactly ONE conversation ({turn} user turns) in JSON format.
- Each turn is from the USER only (no assistant replies).
- The conversation should mix {num_tools} topics from the tool list so that tool selection is challenging.
- The final turn must clearly require ONE of these tools.
- The "answer" field should be the correct tool for the final turn only.
- The "mentioned_tools" field should be a list of ALL tools that appear or are relevant in the conversation (not just the final one).
- Make user utterances natural, conversational, and sometimes misleading by mixing topics.
- The output must strictly follow this JSON format:

{{
    "messages": [
        {{"role": "user", "content": "<user message 1>"}},
        ...
    ],
    "answer": "<one of the tools>",
    "mentioned_tools": ["<Tool1>", "<Tool2>", ...]
}}

Do NOT include any explanations, only return the JSON object.
\end{lstlisting}

\subsection{Entity Extraction}
\textbf{Single-turn prompt.}
\begin{lstlisting}
You are a data generator for a single-turn restaurant reservation entity extraction benchmark.

Your task:
- Generate exactly ONE user message in JSON format (no assistant replies).
- The single user message is about booking a restaurant table and must include:
- Explicit mention of **date, time, and number of people** (all three are required).
    - The date must be expressed in natural language that can be resolved to a fixed calendar date without external knowledge. 
      Acceptable examples include explicit dates ("March 15", "July 4th", "October 21") or fixed holidays such as "New Year's eve", "Christmas" or "Valentine's Day". 
      Do not allow relative references like "this Friday", "next week", or movable holidays like "Thanksgiving".
    - The time must be specific and resolvable (e.g., "7 pm", "noon", "midnight", "9 in the morning"). 
      Avoid vague terms like "afternoon" or "evening".
    - The number of people must be expressed either directly (e.g., "a table for 4") 
      or through unambiguous references (e.g., "my parents and I" -> 3, "the two of us" -> 2). 
      Disallow vague group references such as "my friends" or "a few people".
 - Include unrelated chit-chat (weather, food preferences, travel plans) to make extraction harder.
 - Information can appear in any order within the message.

Return ONLY a valid JSON object in this exact format:

{
  "messages": [
    {"role": "user", "content": "<single user message>"}
  ],
  "answer": {
    "date": "MM/DD",
    "time": "H[am|pm]",
    "people": "N"
  }
}

- The "answer" field must contain ONLY the final reservation details, in the exact format:
- Time uses 12-hour format with "am"/"pm" and hour 1-12 (no leading zero),
- People should be an integer string (e.g., "2", "4").
- The date, time, and people in the "answer" must match the final intended reservation from the message.

Do NOT include any explanations, only return the JSON object.
\end{lstlisting}
\vspace{1em}
\textbf{Multi-turn prompt.}
\begin{lstlisting}
You are a data generator for a restaurant reservation entity extraction benchmark.

Produce exactly ONE conversation in strict JSON.

Global constraints that ALWAYS apply (regardless of types selected):
- Conversation: {turn} turns, and each turn is from the USER only (no assistant lines).
- Explicit mention of **date, time, and number of people** (all three are required).
    - The date must be expressed in natural language that can be resolved to a fixed calendar date without external knowledge. 
      Acceptable examples include explicit dates ("March 15", "July 4th", "October 21") or fixed holidays such as "New Year's eve", "Christmas" or "Valentine's Day". 
      Do not allow relative references like "this Friday", "next week", or movable holidays like "Thanksgiving".
    - The time must be specific and resolvable (e.g., "7 pm", "noon", "midnight", "9 in the morning"). 
      Avoid vague terms like "afternoon" or "evening".
    - The number of people must be expressed either directly (e.g., "a table for 4") 
      or through unambiguous references (e.g., "my parents and I" -> 3, "the two of us" -> 2). 
      Disallow vague group references such as "my friends" or "a few people".
- Vary the order in which date, time, and number of people appear across turns (not always date -> time -> people).
- The final user turn MUST NOT restate all details together in one clean sentence; 
  the final reservation must be inferred by integrating information scattered across turns.
  
Optional feature constraints for THIS sample:
- (change in mind): Include AT LEAST 1~2 changes across date/time/people during the conversation.
- (intermediate chit chat): Sprinkle unrelated chit-chat (weather, preferences, travel) between reservation turns.
- (multiple mention): Occasionally mention multiple reservations for different people, but only ONE final reservation should count in the answer.
\end{lstlisting}


\end{document}